# A Novel Markovian Framework for Integrating Absolute and Relative Ordinal Emotion Information

Jingyao Wu, *Student Member*, *IEEE*, Ting Dang, *Member*, *IEEE*, Vidhyasaharan Sethu, *Member*, *IEEE*, and Eliathamby Ambikairajah, *Senior Member*, *IEEE*

**Abstract**—There is growing interest in affective computing for the representation and prediction of emotions along ordinal scales. However, the term ordinal emotion label has been used to refer to both absolute notions such as low or high arousal, as well as relation notions such as arousal is higher at one instance compared to another. In this paper, we introduce the terminology absolute and relative ordinal labels to make this distinction clear and investigate both with a view to integrate them and exploit their complementary nature. We propose a Markovian framework referred to as Dynamic Ordinal Markov Model (DOMM) that makes use of both absolute and relative ordinal information, to improve speech based ordinal emotion prediction. Finally, the proposed framework is validated on two speech corpora commonly used in affective computing, the RECOLA and the IEMOCAP databases, across a range of system configurations. The results consistently indicate that integrating relative ordinal information improves absolute ordinal emotion prediction.

—————— ◆ ——————

## 1 INTRODUCTION

SPEECH is one of the most natural form of human communication and a key modality through with emotions are expressed. Consequently, speech emotion recognition (SER) has received increasing interest within the speech processing and Human-Computer Interaction research communities [1-4]. Within affective computing systems, emotions are typically represented using either categorical labels or dimensional labels [4-6]. In the former case, emotions are represented as discrete categories using nominal labels (e.g., happy, sad, angry, etc.) [4-6]; and in the latter case, emotion are described using interval scales along affective dimensions such as arousal (activated vs deactivated) and valence (pleasant vs. unpleasant) [7-9]. Additionally, the interval labels can also be used to assign scalar values for arousal and valence at each time step, leading to a continuous time series emotion labels [10, 11]. Both nominal and interval labelling systems have been extensively utilised and investigated. However, the challenge of low inter-rater agreement, i.e. variability in labels reflecting variations in perception among different annotators, remains [11, 12]. This in turn introduces ambiguity in the labels used to train SER systems and increases uncertainty in the predictions [12].

A third framework for labelling which is drawing increasing attention in SER research is referred to ordinal labelling scheme, whereby dimensions such as arousal and valence are still used but instead of interval scales, ordinal scales are used [11, 13]. An ordinal scale is one which defines an order between elements on the scale, but no notion of distance is defined between the elements [13]. For instance, on a scale of low, medium and high, a definite order exists but it is not meaningful to discuss the

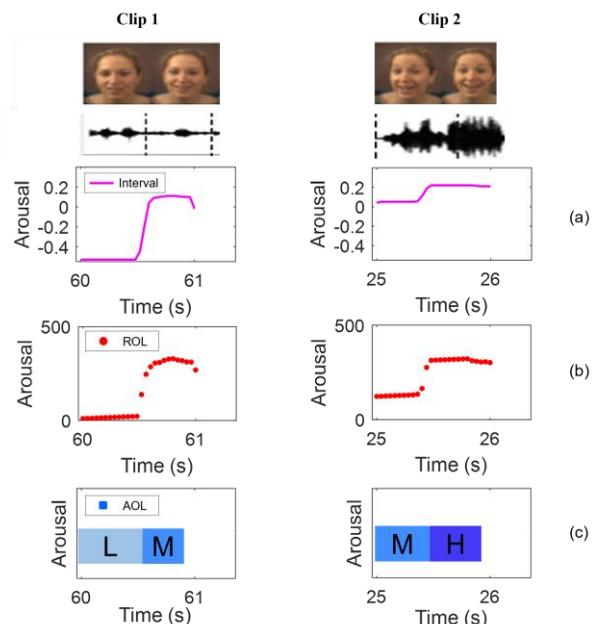

Fig. 1. Two 1-second clips of video and speech recording with attached arousal labels from two utterances. (a) The corresponding annotated interval labels; (b) ROLs indexed at each time instant; and (c) AOLs denoted by different shades of blue indicating three levels: low, medium and high.

- J.Wu, V. Sethu, and E. Ambikairajah are with the School of Electrical Engineering and Telecommunications, University of New South Wales, Sydney, NSW 2052, Australia.
  E-mail: {jingyao.wu, v.sethu, e.ambikairajah}@unsw.edu.au.
- T.Dang is with the Department of Computer Science and Technology, University of Cambridge, Cambridge, UK and also holds a visiting position at the School of Electrical Engineering and Telecommunications, University of New South Wales, Sydney, NSW 2052, Australia
  E-mail: td464@cam.ac.uk

'distance' between the elements. Additionally, research in psychology has shown that people are better at discriminating among options than they are at assigning definitive values for what they perceived [13]. Consequently, several recent studies have advocated that ordinal labelling of emotions is better aligned with human perception, evidenced by lower inter-rater variability [14-18].

While the idea of using ordinal labels is appealing, their use in affective computing systems is fairly recent and the term has been used to refer to two related but distinct types of labels in the emotion recognition literature. In this manuscript, we introduce the terms absolute ordinal labels (AOL) and relative ordinal labels (ROL) to distinguish between these two types of ordinal labels which describe different aspects of emotions. Typically AOLs are obtained by asking annotators to select an element from a finite ordinal scale (e.g., {low arousal, medium arousal, high arousal}) [19], whereas ROLs are obtained by asking annotators to rank the segments within a speech segment from the lowest arousal (or valence) to the highest arousal (or valence) [11]. As illustrated in Fig. 1 (c), AOLs denoted by different shades of blue take values from an 'ordered class', with each label providing an absolute and easily interpretable indication of the affective dimensions at that time (e.g., low arousal). ROLs, denoted by the position of red dots in (b), indicate the rank of that segment with respect to all others within the utterance. These ranks retain the pairwise order relationship between all pairs of segments as inferred from the interval labels shown in (a). ROLs provide a relative label indicating the affect at each time step with respect to other time steps, encapsulating a notion of how emotion changes but lacking an indication of an absolute level.

Both AOLs and ROLs are ordinal but are not equivalent and convey complementary information. Consider two speech segments that are only assigned AOLs – both labelled 'low arousal', even if annotators can determine which of the two corresponds to a lower arousal, the labelling scheme does not encode this. On the other hand, if the two segments are assigned ROLs – for instance, 'A corresponds to lower aroused than B', clearly, there is no indication of the absolute level of arousal and this label cannot distinguish between a scenario where A corresponds to low arousal level and B corresponds to a medium arousal level (Clip 1 in Fig. 1), and one where A corresponds to medium arousal level and B to a high arousal level (Clip 2 in Fig. 1). Current literature in affective computing typically utilises one or another, with both scenarios referred to as 'ordinal labels' [11, 20].

In this paper, we propose a framework that integrates both the absolute and relative ordinal information. We assume that AOLs are more readily interpretable and consequently the predictions of an emotion recognition system should be absolute ordinal quantities. Simultaneously we also recognise that ROLs are better aligned with the types of judgements humans are better at making, and consequently ROL should inform the training and outputs of the emotion prediction system. In the proposed framework, the AOLs are represented as states of a finite state Markov model, with the state transition probabilities predicted by a system trained using ROLs. The overall model, which we refer to as the dynamic ordinal Markov model (DOMM), thus incorporates both information about the emotion at a given time (state probabilities) as well as the change in emotion at a given time (state transition probabilities).

## 2 ORDINAL EMOTION PREDICTION SYSTEMS

In recent years, there has been growing interest in ordinal regression techniques for affective computing and these approaches have shown some advantages in utilising and modelling the ordinal nature of the labels [11]. In this section we summarise recent research on developing both relative and absolute ordinal emotion prediction systems. It should be noted that the terms 'absolute ordinal' and 'relative ordinal' are not used in the literature and the distinction is introduced in this paper.

### 2.1 Relative Ordinal Prediction System

Preference learning (PL), a popular framework for retrieval tasks, refers to a set of ordinal regression models that has become an appealing approach in affective computing. It ranks the samples in ascending/descending order of an emotional attribute. Cao et.al [21] describe the use of rankers for categorical emotion representation with different rankers trained for each emotion category. For instance, a 'happy' ranker trained to treat sample labelled 'happy' with a higher preference than a sample labelled 'sad'. For ranking along emotional attributes or dimensions such as arousal and valence, RankSVMs (a popular PL framework) have been adopted in a number of studies [20, 22-25]. Deep neural network (DNN) based emotional PL algorithms such as RankNet [20] and RBF-ListNet [23] have also drawn increasing attention within the field.

Apart from preference learning methods, ordinal regression tasks have also been framed as binary classification problems, by decomposing the ordinal variables into an ensemble of pairs of classes in a staircase-like manner, leading to the Ordinal Binary Decomposition Approach [26]. Finally, naïve approaches which involve treating ordinal labels as interval or nominal labels (and ignoring their ordinal nature) have also been investigated. For instance, a conventional regression model can be trained by treating the relative ordinal labels $[r_1, ..., r_N]$ as $N$ different interval values $[1, 2, ..., n, ..., N]$ (e.g., assign the rank order as real numerical values, $r_n = n$) [26, 27]. However, as stated in [26, 28] the regressor will be more sensitive to the converted values rather than the original rank order, which might degrade the performance of the regression models.

### 2.2 Absolute Ordinal Prediction System

AOLs have been adopted widely in current emotion recognition studies, but they are generally treated as nominal labels and the underlying ordinality is ignored. For instance, in a few different studies [29-32], affect labels in terms of low, medium and high arousal are utilised, but naïve nominal classifiers, such as support vector machine (SVM), are employed [30].

An alternative approach is the threshold method that assumes a continuous latent variable represents the emotional state, with appropriate thresholds dividing the range of the latent variable into a few ordered ranges corresponding to absolute ordinal labels. The prediction model then learns a mapping function between the input features and the continuous hidden latent variable, which is then converted to an absolute ordinal label. Several modelling methods can be utilised in this approach, such as cumulative link models [33], and proportional odds models [34].

Amongst truly ordinal prediction systems, the Ordinal Multi-class SVM (OMSVM) which was first proposed by Kim and Ahn [35] for credit ratings and has since been widely used in the area of financial and market analysis [36, 37], appears to be a very promising candidate. Additionally, a suitable ordinal binary decomposition approach can also be employed to model AOLs. Compared to the naïve approach and threshold methods, the binary decomposition method is more suitable since it can capture ordinal information without defining distances between ordered classes. Deep learning may also be a valid solutions to ordinal binary decomposition, but it can suffer with generalisation, especially for small dataset [35, 38].

## 3 PROPOSED DYNAMIC ORDINAL MARKOV MODEL

### 3.1 Combining Absolute and Relative Ordinal Information

The task of integrating two aspects of ordinal emotion labels with AOL reflecting the emotional state (static) and ROL indicating change in emotional state (dynamic) can be intuitively associated with a Markovian framework. A Markov model (MM) is a stochastic model describing a sequence of states, where the probability of occurrence of future states depends only on the current state [39]. In the context of ordinal emotion labels, AOL and ROL can be viewed as informing the state and transition probabilities of a MM. For instance, it is reasonable to expect that a positive change in the ROL (increase in rank) would indicate high probabilities for transitions from a 'low' arousal/valence state to a 'medium' arousal/valence state as well as 'medium' to 'high', while simultaneously indicating a low probability for other transitions (like 'high' to 'medium', 'high' to 'low', etc.). Similarly, a large negative change in ROL would indicate a high probability for a 'High' to 'Low' transition and a low probability for all other transitions.

This idea is illustrated in Fig. 2, with the position of the blue circles indicating the AOL with $A_0$, $A_1$, and $A_2$ denoting low (L), medium (M) and high (H) arousal states respectively; and the values within the circles denoting the ROL (rank within a set of 100 time steps). The number on each arrow given by Δ shows the change in ROL between two consecutive instances. It is clear that the value of Δ carries information about the dynamics of the AOL, i.e., any change in $A_i$. For example, the chances of erroneously predicting the AOL at time $t_4$ as low or medium instead of high would be greatly reduced if the automatic system took into account that over the $t_{3\to4}$ transition, ROL change from rank 48 to 87 (i.e., Δ = 39).

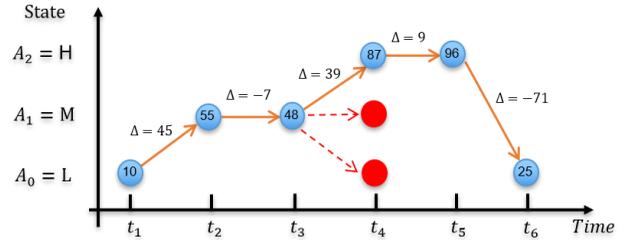

Fig. 2. A graphical representation of the complementary nature of absolute and relative ordinal labels over 6 time steps (within a 100 time step utterance). The position of the blue circles represents AOLs at different time steps with ROLs indicated by the number insides the circles, and the arrows depicting ROL changes between consecutive time steps. The red circles represent the potential AOLs at time $t_4$ but would be ruled out given the Δ over the $t_{3\to4}$ transition.

An overview of the proposed Dynamic Ordinal Markov Model (DOMM) based emotion prediction system is shown in Fig. 3. It comprises two subsystems which operate in parallel, one to predict AOLs and one to predict ROLs from the input features. The AOL prediction subsystem is designed to estimate state posteriors, $P(\beta_t|x_t)$, given the input features, $x_t$, at each time step $t$. The possible states of the model, $\beta$, correspond to the finite set of possible AOLs. In all the systems presented in this paper, there are possible states – 'low', 'medium' and 'high' for arousal or valence.

The output of the ROL prediction subsystem is a rank, $\alpha_t$, at each time step $t$; from which state transition probabilities, $P(\beta_t|\beta_{t-1}, \Delta\alpha_t)$, are inferred based on the rank difference between consecutive time steps, $\Delta\alpha_t = \alpha_t - \alpha_{t-1}$ as per Bayes theorem:

$$P(\beta_t | \beta_{t-1}, \Delta\alpha_t) = \frac{P(\Delta\alpha_t | \beta_{t-1}, \beta_t) P(\beta_t | \beta_{t-1})}{P(\Delta\alpha_t | \beta_{t-1})} \quad (1)$$

where $P(\Delta\alpha_t | \beta_{t-1}, \beta_t)$ represents the probability of rank (ROL) difference $\Delta\alpha_t$ given the state (AOL) at previous time step, $\beta_{t-1}$, and the current state, $\beta_t$; $P(\Delta\alpha_t | \beta_{t-1})$ denotes the probability of a change in rank (ROL) of $\Delta\alpha_t$ occurring given previous state was $\beta_{t-1}$; and $P(\beta_t | \beta_{t-1})$ denotes the probability of transitioning from state $\beta_{t-1}$ to state $\beta_t$. Models for all three probabilities on the right-hand side of (1) can be inferred from labelled training data.

To model $P(\Delta\alpha_t | \beta_{t-1})$, first all instances of ROL pairs [$\alpha_t$, $\alpha_{t-1}$] with initial state $\beta_{t-1} = L$ from the training data are selected and the corresponding set of $\Delta\alpha_t$ values are computed. Kernel density estimation (KDE) [40, 41] is then carried out to obtain a model for $P(\Delta\alpha_t|\beta_{t-1} = L)$. Similarly models for $P(\Delta\alpha_t|\beta_{t-1} = M)$ and $P(\Delta\alpha_t|\beta_{t-1} = H)$ are obtained.

Models for $P(\Delta\alpha_t|\beta_{t-1}, \beta_t)$ can also be obtained by the same approach, by partitioning the dataset into 9 subsets corresponding to all combinations of $\beta_{t-1}, \beta_t \in \{L, M, H\} \times \{L, M, H\}$, following by KDE. An example is illustrated in Fig. 4 with: Fig. 4(a) showing the histogram of $\Delta\alpha_t$ for low-low (L-L) transitions (from the RECOLA dataset); Fig. 4(b) showing the histogram for high-low (H-

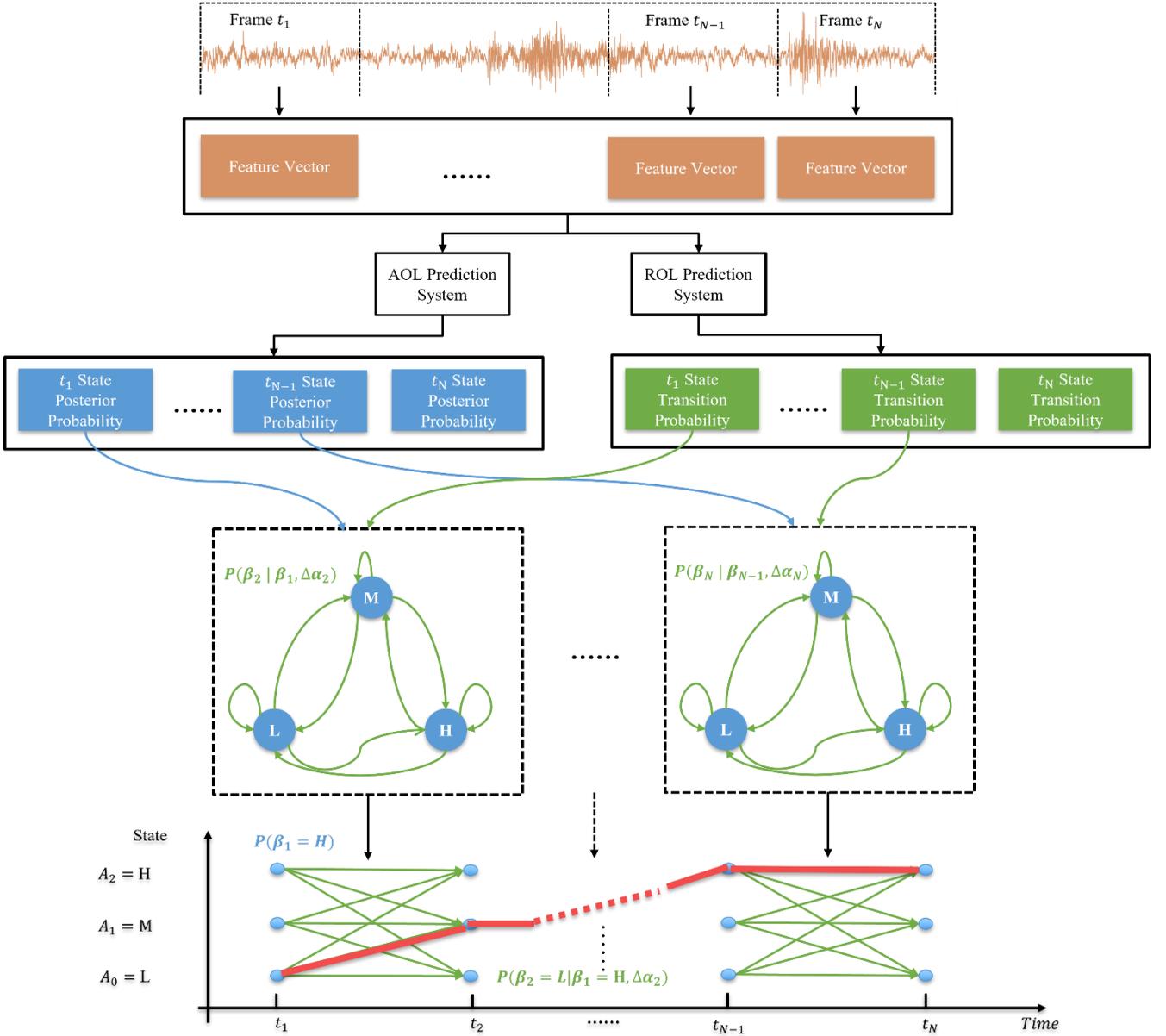

Fig. 3. Overview of the proposed DOMM system architecture. A graphical representation of the Markov model diagram is shown in the middle with blue circles representing the system state $\beta_t = \{L, M, H\}$ at time $t$ and green arrows denoting transition probabilities $P(\beta_t \mid \beta_{t-1}, \Delta\alpha_t)$. Both of these parameters change with time and inferred based on the AOL and ROL prediction systems, making the model dynamic. A representation of the state lattice is given at the bottom, where the y axis $A_i, i = [0,1,2]$ represents the AOLs comprising of low(L), medium (M) and high (H) states, and the x axis represents the time step from $t_1$ to $t_N$. Blue dots corresponding to the state posterior probabilities calculated from AOL prediction system at each time instant. The transition probabilities between each state between each pair of time frames (green arrow). Red line indicating the 'best path' after Viterbi decoding which gives the final predicted AOL sequence.

L) transitions; and Fig. 4(c) showing the models for $P(\Delta\alpha_t \mid \beta_{t-1} = L, \beta_t = L)$ and $P(\Delta\alpha_t \mid \beta_{t-1} = H, \beta_t = L)$ obtained via KDE. It can be seen that $\Delta\alpha_t$ is much more likely to take small positive or negative values given L-L transitions, while it is much more likely to take large negative values given H-L transitions.

Lastly, the probability $P(\beta_t \mid \beta_{t-1})$ is calculated as:

$$P(\beta_t = j \mid \beta_{t-1} = i) = \frac{N_{i \to j}}{N_i} \quad (2)$$

where, $i, j \in \{L, M, H\}$; $N_i$ denotes the number of instances of state (AOL) $i$ in the training data set; and $N_{i \to j}$ denotes the number of instances of state $i$ followed by state $j$ in the training set.

Together, the state posteriors, $P(\beta_t \mid x_t)$, and the state transition probabilities, $P(\beta_t \mid \beta_{t-1}, \Delta\alpha_t)$, describe a 'dynamic' ordinal Markov model (DOMM). Here the term 'ordinal' refers to the fact that the model represents ordinal labels and the term 'dynamic' refers to the fact that both the state posterior probabilities and the state transition probabilities are time-varying, with both dependent on the input features, $x_t$, at time $t$. Note that $\Delta\alpha_t$ is estimated from $x_t$ by the ROL prediction subsystem and hence the state transition probabilities are also dynamically inferred from the input features. Given a sequence of input features, $\{x_t; t_1 \le t \le t_N\}$, the DOMM generates a state lattice and the most probable sequence of states (ab-solute ordinal labels) can be inferred using the Viterbi algorithm [42].

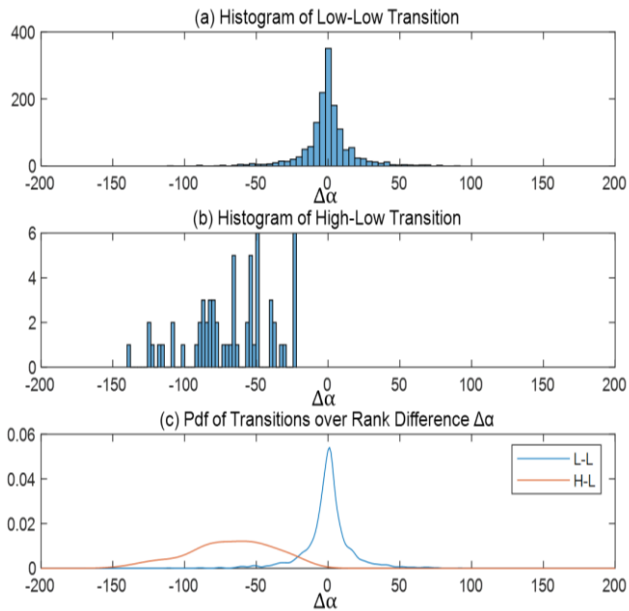

Fig. 4. Example of histograms and PDF distributions of different state transitions. Top: histogram of low-to-low transition. Middle: Histogram of high-to-low transition. Bottom: Pdf $P(\Delta\alpha_t \mid \beta_{t-1} = i, \beta_t = j)$ of the two transitions on the same plots. Y-axis of the top two histogram indicates the number of frame pairs, and that for the bottom pdf indicate the PDF value. X-axis of three sub-plots represents possible values of ROL difference.

## 3.2 Subsystem Descriptions

Any appropriate machine learning model can be used as the AOL and ROL prediction subsystems in the proposed framework. In the experiments reported in this paper, we use an Ordinal Multiclass Support Vector Machine (OMSVM) as the AOL prediction subsystem and a RankSVM based model as the ROL prediction subsystem. They have both been shown to be robust to overfitting and their mathematical framework is well understood [43, 44]. Furthermore, both explicitly model the inherent ordinal nature of the labels [35, 45]. An overview of both subsystems is provided below.

### 3.2.1 AOL Prediction Subsystem: OMSVM

OMSVM [35, 38] is a variation of the conventional Multi-class SVM (MSVM) that makes use of the additional structure imparted to the data by the ordinal nature of the labels [46-49]. Instead of treating the ordinal labels as independent classes, as would be the case with conventional MSVM systems, OMSVM first applies an ordinal pairwise partitioning (OPP) technique that divide the ordinal labels into groups as OneVsNext: $O_1\&O_2$, $O_2\&O_3$…, and then uses a series of SVMs for each group. During the prediction (test) phase, the test data is passing through the SVM classifiers in order, either along the forward or backward direction, until a class is predicted, at which point the remaining SVMs will not be used. To convert the outputs of OMSVM into state posteriors, we fit a sigmoid function to the outputs as suggested in [50]:

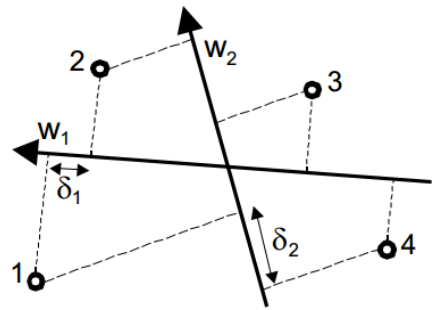

Fig. 5. Points in feature space are ranked by two weight vectors. The projection of points on the vector indicates ranks [45]. For instance, for hyperplane $\vec{w}_1$, the rank is 1≻2≻3≻4, while for $\vec{w}_2$ the rank is 2≻3≻1≻4.

$$P(\beta_t = i \mid y) = \frac{1}{1 + exp(ay + b)} \quad (3)$$

where $y$ denotes the SVM output; and $a$ and $b$ refer to sigmoid function parameters which are determined during training as outlined in [50]. For further details, readers are referred to [46, 50].

### 3.2.2 ROL Prediction Subsystem: RankSVM

Conventional SVM classifiers aims to find a hyperplane $\vec{w}$ that maximally separates two classes, and RankSVMs [45] extend this idea to identify a hyperplane that performs so-called preference comparisons, such that projections of the data on that hyperplane still preserve the original rank orders. This idea is illustrated in Fig. 5, where it can be seen that the optimal hyperplane would be $w_1$ (instead of $w_2$) since the projections onto the hyperplane corresponding to ranks (ROLs) $1 \succ 2 \succ 3 \succ 4$ along the hyperplane (where '≻' denotes 'preference') still preserve its order.

## 4 DATABASE DESCRIPTION

None of the publicly available speech corpora come with absolute and relative ordinal labels. Consequently, in this work we use the well-established RECOLA [51] and the IEMOCAP [52] databases and convert the interval labels to AOL and ROL.

### 4.1 RECOLA

The Remote Collaborative and Affective Interactions (RECOLA) dataset [51] is a widely used multimodal corpus containing both audio and video modalities. It consists of 23 dyadic interactions from 46 participants including 27 females and 19 males. The data used in the experiments reported in this paper corresponds to that used in the AVEC challenge 2016 [53, 54], which consists of 9 utterances each in the training and development sets (with no overlap) with each utterance having a duration of 5 minutes. Results on the development set are reported as test set labels have not been publicly released. Each utterance is also annotated by 6 raters with continuous arousal and valence ratings between -1 to 1 at a sampling interval of 40ms.

Delay compensation is applied to compensate the human perception delays in the labels as suggested in [55]

with 4 seconds for arousal and 2 seconds for valence. Additionally, the labels are smoothed by averaging within 1sec windows, with 50% overlap between windows. This is in line with the suggested use of 1-3 second windows to capture the trend in the interval labels [56]. In total, each utterance (5 minute duration) comprised of 615 and 617 windows for arousal and valence.

### 4.1.1 Interval to AOL Conversion:

Absolute ordinal labels can be obtained by thresholding the interval labels into three ordinal levels - low, medium, and high. However, choosing these thresholds are a challenge and no commonly agreed upon thresholds exist [29, 57]. Therefore, a range of potentially suitable thresholds were analysed. Additionally, the conversion is carried out individually for each annotator and the final consensus AOL is inferred via majority vote among the 6 individual AOLs (one per annotator).

Given the label value $\bar{y}_t$ representing the average arousal/valence intensity within each window and thresholds $\theta_1$ and $\theta_2$, the AOL is taken as Low, if $\bar{y}_t \leq \theta_1$; Medium, if $\theta_1 < \bar{y}_t \leq \theta_2$; and High, if $\bar{y}_t > \theta_2$. To determine suitable values for $\theta_1$ and $\theta_2$, measures of label balance and inter-rater agreement after conversion are analysed. Label balance ($\gamma$) is computed based on the distribution of three classes occurred within training set. First, the difference in relative frequency between the most and least frequent AOLs within training dataset is computed as $\gamma = \left|\frac{N_m - N_l}{N}\right|$, where $N_m$ and $N_l$ represent number of most and least frequent AOLs respectively, and $N$ is the total number of AOLs within the training set. A large $\gamma$ indicates highly imbalanced label distribution and vice versa. Inter-rater agreement represents consensus level among the converted AOL across multiple annotators. It is quantified as the ratio of the number of frames where more than half of the raters agree on the same AOL. A reasonably high ratio means a higher inter-rater agreement.

There will be a trade-off between the label balance and level of agreement. For instance, thresholding all annotations into a single AOL leads to perfect agreement but also a highly skewed and uninformative label distribution. Whereas, equally spacing the thresholds is likely to lead to less distinguishable labels that are unlikely to be optimal except in the case of uniform distribution of interval labels, at unrealistic expectation – extreme affect labels are less frequent.

Finally, as arousal annotations were observed to be reasonably symmetrically distributed over [-1, 1], the thresholds $\theta_{a1}$ and $\theta_{a2}$ for arousal were also set to be symmetric, with $\theta_{a2} = -\theta_{a1}$. A range of different values of $\theta_{a2}$, within [0.08, 0.2] with a step size 0.02, was analysed. This region was selected based on the observation that the median arousal value from every annotator fell in the range $[-0.067, 0.118]$ and we wanted to ensure that the 'medium' label always covers the median values. Both the label balance measure and the inter-rater agreement for the chosen range of thresholds are shown in Fig. 6(a).

It was also observed that valence labels in RECOLA

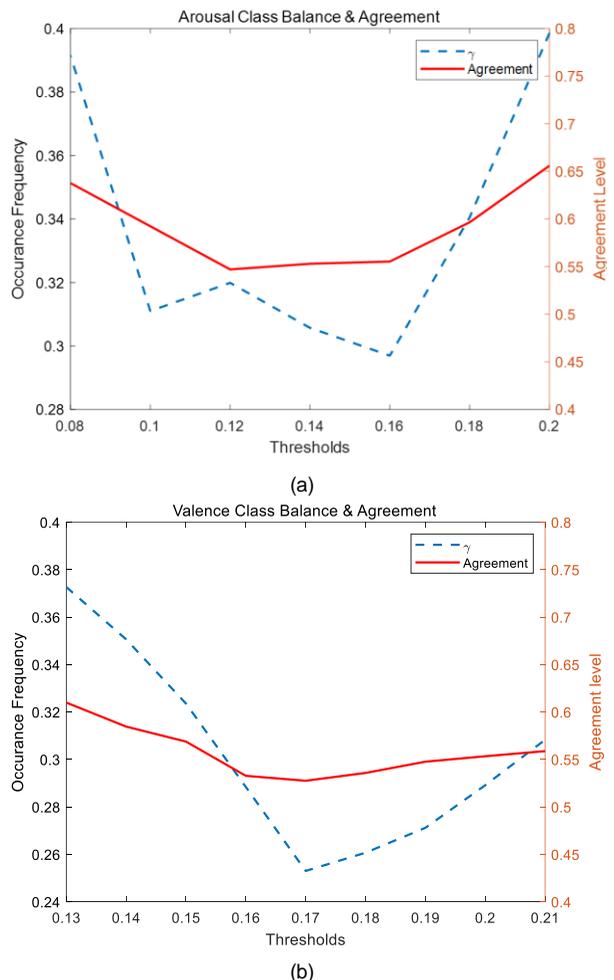

Fig. 6. Key characteristics (label balance & agreement) of converted labels for different thresholds on (a) arousal and (b) valence. X-axis denotes the $\theta_2$ values.

were skewed towards positive values and consequently the two thresholds, $\theta_{v1}$ and $\theta_{v2}$, were assumed to be symmetrical about 0.085 and instead analysed over the range $\theta_{v2} \in [0.13, 0.21]$ with $\theta_{v1} = 0.17 - \theta_{v2}$. These ranges were also chosen so as to ensure the median label value from each annotator was covered by the 'medium' label. The label balance measure and inter-rater agreement variations across the range of $\theta_{v2}$ is depicted in Fig. 6(b).

In Fig. 6, the. x-axis denotes the upper threshold $\theta_{a2}$ and $\theta_{v2}$ values, the left y-axis denotes the label balance and right y-axis denotes the inter-rater agreement levels. Label balance is plotted as the average $\gamma$ across all annotators (i.e., $\bar{\gamma}$) and represented by a blue dashed line, and the inter-rater agreement level is indicated by the solid red line.

It can be seen from Fig. 6(a) that $\bar{\gamma}$ tends to converge to the minima within the range $\theta_{a2} \in [0.14, 0.18]$, suggesting this would lead to the most balanced label distribution. The agreement level within this range is between 0.54 to 0.6 which is relatively high as well. It should also be noted that the higher agreement level outside this range for $\theta_{a2}$ mostly likely arises from imbalanced labels – i.e., a large number of instances are assigned to one label. Similar finding were observed for valence, and threshold val-ues ranging around $\theta_{v2} \in [0.16, 0.18]$ were found to be suitable. Finally, a different set of AOLs for different thresholds within the range $\theta_{a2} \in [0.12, 0.18]$ for arousal; and $\theta_{v2} \in [0.14, 0.18]$ for valence with step size 0.01 were inferred and all experiments were repeated using each set and the mean and standard deviation across them are reported in this paper.

### 4.1.2 Interval to ROL Conversion

ROLs are similarly inferred from smoothed interval labels, obtained by first windowing the interval labels with 1 sec windows (with 50% overlap) and computing the mean within each window. A rank sequence of ROLs is then obtained by performing pairwise comparisons for each individual annotator, and global ROLs were inferred using the Qualitative Agreement (QA) method across all annotators [58]. Within each utterance, a matrix of pairwise comparisons amongst all windows for each individual annotator is first collected as shown in Fig. 7. A consensus matrix is then obtained via majority vote among matrices from all annotators and the final rank sequence of ROLs is obtained from this consensus matrix.

## 4.2 IEMOCAP

The Interactive Emotional dyadic MOtion CAPture database (IEMOCAP) [52] contains 12 hours of audio-visual recordings of 5 dyadic sessions from 5 pairs of actors. Both scripted dialogs and improvised dialogs are collected, and each dialog is segmented into speaker turns. In total the corpus contains 10,039 turns with an average duration of 4.5 s. IEMOCAP is annotated at a turn level that each turn is attached with both a categorical emotion label (i.e., happiness, anger, sadness, neutral state and frustration) and primitive based interval annotations (valence, activation, and dominance) on a 5-point scale (i.e., 1 -low/negative, 5 – high/positive). However, only the interval labels along arousal and valence within the range [1,5] are considered in this work and the experiments on IEMOCAP are carried out at the turn-level with the same leave-one-speaker-out cross validation as in [29].

A number of different conversion schemes has been employed to convert interval labels in IEMOCAP to AOLs, but they lack a consistent set of thresholds [29, 30, 59-61]. Therefore, we adopted the idea of clustering the labels in the dataset to identify suitable decision thresholds [59, 61].

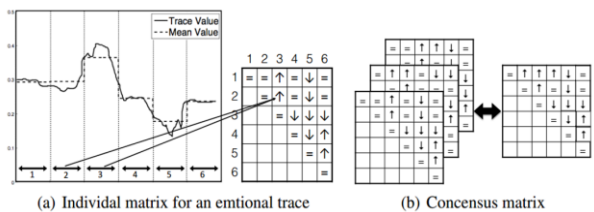

Fig. 7. Illustration of QA method [58]. (a) Individual comparison matrix obtained from interval labels for one rater. Up-arrow indicates an increase; down-arrow indicates a decrease and equal denotes tie. (b) Consensus matrix obtained by aggregating individual matrix collected from multiple raters using majority votes.

TABLE 1
INTERVAL TO ABSOLUTE ORDINAL LABEL CONVERSIONS.

| | Arousal | | |
|---|---|---|---|
| | Low | Medium | High |
| IEMOCAP | [1, 2.5) | [2.5, 3.5) | [3.5, 5] |
| RECOLA | [-1, $\theta_{a1}$] | ($\theta_{a1}$, $\theta_{a2}$) | [$\theta_{a2}$, 1] |
| $\theta_{a1} \in [-0.12, -0.18]$ $\theta_{a2} \in [0.12, 0.18]$ | | | |

| | Valence | | |
|---|---|---|---|
| | Low | Medium | High |
| IEMOCAP | [1, 2.8] | (2.8, 4) | [4, 5] |
| RECOLA | [-1, $\theta_{v1}$] | ($\theta_{v1}$, $\theta_{v2}$) | [$\theta_{v2}$, 1] |
| $\theta_{v1} \in [-0.03, 0.01]$ $\theta_{v2} \in [0.14, 0.18]$ | | | |

TABLE 2
ABSOLUTE ORDINAL LABELS DISTRIBUTION

| | IEMOCAP | | | |
|---|---|---|---|---|
| | | Low | Medium | High |
| Training set | Arousal | 932±42 | 2180±40 | 1319±51 |
| | Valence | 1310±26 | 1534±52 | 1580±34 |
| Testing set | Arousal | 103±42 | 242±40 | 146±51 |
| | valence | 149±26 | 170±52 | 176±34 |

| | RECOLA | | | |
|---|---|---|---|---|
| | | Low | Medium | High |
| Training set | Arousal | 1555±274 | 1730±948 | 2250±674 |
| | Valence | 1814±454 | 2002±190 | 1754±327 |
| Testing set | Arousal | 2424±255 | 1335±736 | 1775±483 |
| | Valence | 2175±428 | 1868±162 | 1528±372 |

Using K-means clustering, this led to the following thresholds for arousal: $Low \in [1, 2.5)$, $Medium \in [2.5, 3.5)$, and $High \in [3.5, 5]$; and these for valence: $Low \in [1, 2.8)$, $Medium \in [2.8, 4)$, and $High \in [4, 5]$. Interval to ROL conversion in the IEMOCAP data was carried out in a similar manner to that in RECOLA, but based on pairwise comparisons between speaker turns instead of 1 second windows [20, 22].

The label conversions threshold adopted for both the IEMOCAP and RECOLA datasets are summarized in Table 1 and Table 2 shows the label distributions obtained after conversion for the training and test sets. Specifically, all the experimental results reported in this paper using the IEMOCAP database, the experiments are carried out-with leave-one-speaker-out cross validation resulting in 10-fold cross validation. Consequently, the mean and standard deviation of number of turns used in training and testing set across the 10 folds are reported in Table 2. Simiarly, all experiments on the RECOLA database, was conducted for different threshold settings, which results in different numbers for each class. The mean and standard deviation of samples across all thresholds is reported.

# 5 EXPERIMENT SETTINGS

## 5.1 Features

The 88-dimensioanl extended Geneva Minimalistic Acoustic Parameter Set (eGeMAPS), a relatively standard feature set used in affective computing to simplify benchmarking, was adopted for all the experiments reported in this paper [62, 63]. The feature set comprises of arithmetic mean and coefficient of variation functionals applied to 18 low-level descriptors (LLDs) extracted from the minimalistic acoustic parameter set along with another 8 functionals applied to pitch and loudness. Additional 7 LLDs are extracted from the extension parameter set with 4 statistics over the unvoiced segments, 6 temporal features, and 26 additional cepstral parameters and dynamic parameter [62]. The features were extracted using the OpenSMILE toolkit [64] and for additional details about eGeMAPS, readers are referred to [62].

## 5.2 System Parameters Settings

As previously mentioned, the AOL prediction subsystem in the proposed framework was implemented as an ordinal multiclass SVM (OMSVM) in the reported experiments. This OMSVM implementation used the *ClassificationECOC* MATLAB toolbox (an error correction output code multi-class classifier) [65]. The state posterior probabilities were then computed using the *FitPosterior* function [50].

The RankSVM model used in the ROL prediction subsystem was implemented using the toolkit referred to in [66]. It is trained using primal Newton method, which is known to be fast, [67, 68] and the maximum Newton step was set to be 20 as in [66]. Finally, both the OMSVM and RankSVM models utilised linear kernels and both used $c = 1 \times 10^{-4}$ as suggested in [30, 69].

## 5.3 Evaluation Metrics

### 5.3.1 AOL Prediction Evaluation Metrics

Unweighted Average Recall (UAR) is a standard metric used to quantify performance in nominal classification tasks [29, 30], and has also been utilised to report 'classification accuracy' even predicting AOLs [35]. We report UAR as one measure to allow comparison with existing literature. However, we also note that UAR does not take into account ordinality in the labels. For instance, incorrectly predicting 'Low arousal' as 'Medium arousal' or 'High arousal' both carry the same penalty although obviously the latter is a bigger error. Therefore, to take the ordinal nature of AOLs into consideration, we also report the weighted Cohen's Kappa (WK) coefficient, $k_w$, which is used to measure the consistency between two AOL sequences. The coefficient, $k_w$, indicates the level of agreement between two different label sequences (predictions Vs ground truth) as given by (4), with $k_w = 1$ indicating perfect agreement and $k_w = 0$ indicating only chance agreement [70].

$$k_w = \frac{\sum_{i=1}^{3}\sum_{j=1}^{3} w_{ij}\, p_{ij} - \sum_{i=1}^{3}\sum_{j=1}^{3} w_{ij}\, p_{i.}q_{.j}}{1 - \sum_{i=1}^{3}\sum_{j=1}^{3} w_{ij}\, p_{i.}q_{.j}} \quad (4)$$

$$W = \begin{bmatrix} 1 & 0.5 & 0 \\ 0.5 & 1 & 0.5 \\ 0 & 0.5 & 1 \end{bmatrix} \quad (5)$$

Where $p_{ij} = \frac{n_{ij}}{N}$ denotes the proportion of each element of the 3 × 3 confusion matrix that maps the ground truth and prediction AOLs, $i,j = [1,2,3]$ referred to AOL - low, medium, high. $p_{i.} = \frac{n_{i.}}{N}$ denotes the proportion of AOL $i$ of the ground truth with $n_{i.}$ referred to the number of AOLs equal to $i$. $q_{.j} = \frac{n_{.j}}{N}$ denotes the proporrtion of AOL $j$ of the predictions with $n_{.j}$ referred to the number of AOLs equal to $j$. $N$ denotes the total number of AOLs and $w_{ij}$ is the element of $i^{th}$ row and $j^{th}$ column of matrix $W$ defined in (5).

### 5.3.2 ROL Prediction Evaluation Metrics

To quantify the performance of the ROL prediction subsystem we adopt measure called precision at k (P@k), which is widely used in the rank-based retrieval tasks and in preference learning tasks [20, 22]. It is a measurement of precision in retrieving top or bottom k% of the samples. It is calculated by first dividing ground truth labels into high and low groups following the rule that ROLs larger than median is collected in the high group and vice versa as shown in top section of Fig. 8. The prediction is counted as a success if the top $k$% of the predictions belong to high group and the bottom $k$% samples of the predictions are retrieved from the low group as depicted in Fig. 8.

The other commonly used metric for quantifying mismatch between two ranks is Kendell's Tau [71]. It ranges from -1 to 1 indicating the range from completely antithesis to perfect matched, with 0 indicating the two ranks are independent [72]. Kendall's Tau, $\tau$, is given as:

$$\tau = \frac{C - D}{T} \quad (6)$$

where $T$ refers to total number of comparisons given by $T = \frac{n(n-1)}{2}$, with $n$ referring to the highest rank index. $C$ denotes the number of concordant pairs and $D$ denotes the number of discordant pairs.

It should be noted that both P@k and $\tau$ are used to evaluate a single sequence of ROLs. In the reported experiments, P@k and $\tau$ are first calculated for each utterance and the mean over all the utterances in the test set is then reported.

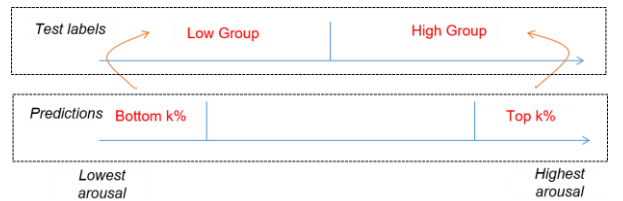

Fig. 8. A graphical representation of P@k. The test labels are sorted and divided into low/high groups. If the top k% of the predictions are from High groups, this is success, same for bottom k%

# 6 EXPERIMENTAL RESULTS AND DISCUSSION

Prior to testing the proposed DOMM based prediction system, we quantify the performance of the AOL and ROL prediction susbsystems individually based on the metrics outlined in the previous section. Experimental results on the IEMOCAP database are reported in terms of the mean and standard deviation computed across each fold of the 10-fold cross validation as discussed in section 4. Results on the RECOLA database are reported in terms of means and standard deviations of the performance metrics determined on the different AOL sets obtained using the identified range of thresholds ($\theta_{a2}$ and $\theta_{v2}$) for the interval label to AOL conversions (refer section 4.1).

## 6.1 OMSVM Performance

The performance of the OMSVM based AOL prediction subsystem evaluated on both IEMOCAP and REOCLA in terms of UAR are reported in Table 3. On the IEMOCAP database, the OMSVM system is compared to an HMM based baseline [29]. While the UAR of the OMSVM system is lower than that of the baseline, it is reasonably high and significantly higher than chance. Consequently, we expect the OMSVM system to constitute a reasonable AOL prediction subsystem to evaluate the proposed framework. Finally, it should be noted that the baseline system results were reported with a different AOL conversion scheme.

No suitable baseline systems that allow for a direct comparison on the RECOLA database could be identified since most published literature using RECOLA deals with interval label prediction. The closest reported approaches focus on binary classification tasks of Low/High arousal/valence [31, 32]. Thus, the binary classification [31] was adopted as a reference system (refer Table 3). The performance of the OMSVM subsystem is lower than that of reference system, but the OMSVM system's target is one of three AOLs while the reference system is solving a two-class problem.

TABLE 3
UNWEIGHTED AVERAGE RECALL (UAR) AND WEIGHTED KAPPPA ($k_w$) MEASUREMENT OF OMSVM SUBSYSTEM.

| | IEMOCAP | | | |
|---|---|---|---|---|
| | **Arousal** | | **Valence** | |
| | UAR (%) | $k_w$ | UAR (%) | $k_w$ |
| OMSVM | 53.1±3.79 | 0.279±0.06 | 46.3±4.67 | 0.190±0.06 |
| HMM [29] | 61.9±4.88 | - | 49.9±3.63 | - |
| | **RECOLA** | | | |
| | **Arousal** | | **Valence** | |
| | UAR (%) | $k_w$ | UAR (%) | $k_w$ |
| OMSVM | 55.3±2.08 | 0.443±0.05 | 36.4±1.77 | 0.079±0.03 |
| Binary Classification [31] | 60.7 | - | 52.3 | - |

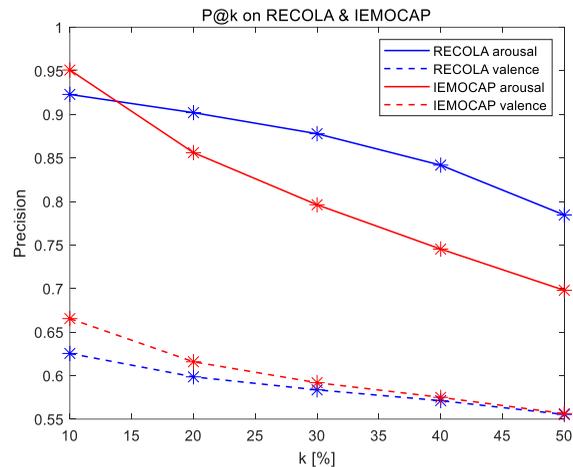

Fig. 9. P@k on RECOLA and IEMOCAP at k values: P@10, P@20 P@30, P@40 and P@50.

TABLE 4
KENDALL'S TAU MEASUREMENTS OF RANKSVM.

| | **RECOLA** | **IEMOCAP** |
|---|---|---|
| **Arousal** | 0.500 | 0.371±0.06 |
| **Valence** | 0.130 | 0.064±0.06 |

## 6.2 RankSVM Perforamce

The performance of the RankSVM based ROL prediction subsystem is quantified in terms of both Kendall's Tau and P@k. Fig. 9 shows P@k for k = 10%, 20%, 30%, 40% and 50%. The solid lines correspond to arousal and dashed lines to valence with the red and blue colours representing performance on IEMOCAP and RECOLA respectively. On the IEMOCAP dataset, the RankSVM system achieves a P@k of around 95% and 67% for arousal and valence at k=10% which is better than the 85% and 63% reported in [20].

Additionally, recognising that P@k does not reflect the exact ranking order within the samples but only indicates the performance of retrieving top and bottom k% of samples, we also report Kendall's Tau in Table 4. The performance of the RankSVM system for valence prediction is uniformly poorer than arousal rank prediction and while this is consistent with reported results on interval labels as well when using audio inputs [54, 73], the Kendall's Tau value for valence rank prediction in IEMOCAP is close to chance.

## 6.3 Evaluating the proposed DOMM framework

The DOMM framework was developed to integrate complementary information from independent AOL and ROL prediction subsystems. Noting that the proposed approach ultimately predicts AOLs, UAR and weighted Cohen's kappa ($k_w$) are employed as performance metrics. Also recall that UAR does not take the ordinal nature of the labels into account, while $k_w$ does. The proposed system, denoted as DOMM_RS, is compared to the basic OMSVM system that does not make use of any relative ordinal information. In addition, we also report the per-

TABLE 5
UNWEIGHTED AVERAGE RECALL (UAR) AND WEIGHTED KAPPA ($k_w$) - MEASUREMENT IN TERMS OF MEAN AND STANDARD DEVIATION OF 10-FOLD CROSS-VALIDATION ON IEMOCAP, AND DIFFERENT THRESHOLDS ON RECOLA.

**IEMOCAP**

|  | Arousal | | Valence | |
| --- | --- | --- | --- | --- |
|  | UAR (%) | $k_w$ | UAR (%) | $k_w$ |
| OMSVM | 53.1±3.79 | 0.279±0.09 | 46.3±4.67 | 0.190±0.06 |
| DOMM_RS | 53.2±4.98 | 0.252±0.11 | 51.1±4.94 | 0.232±0.07 |
| DOMM_GT | 60.9±5.10 | 0.363±0.12 | 59.2±7.13 | 0.358±0.10 |
| HMM [36] | 61.9±4.88 | - | 49.9±3.63 | - |

**RECOLA**

|  | Arousal | | Valence | |
| --- | --- | --- | --- | --- |
|  | UAR (%) | $k_w$ | UAR (%) | $k_w$ |
| OMSVM | 55.3±2.08 | 0.443±0.05 | 36.4±1.77 | 0.079±0.03 |
| DOMM_RS | 55.9±2.14 | 0.462±0.07 | 38.4±2.07 | 0.113±0.04 |
| DOMM_GT | 57.7±2.52 | 0.493±0.06 | 40.8±1.88 | 0.168±0.03 |

formance of a version of the DOMM framework that uses rank difference ($\Delta\alpha_t$) from the ground truth ROLs instead of the ROLs predicted by the RankSVM system. This system, denoted as DOMM_GT, serves as an indication of the upper bound of what is possible when integrating relative and absolute ordinal information with the DOMM framework. The results from both IEMOCAP and RECOLA are reported in Table 5. It can be seen from these results that consistently, across both databases and across both arousal and valence, the DOMM_RS outperforms the OMSVM system and is in turn outperformed by the DOMM_GT system. The consistent trends strongly support the idea that absolute and relative ordinal labels contain complementary information and jointly modelling them has clear benefits.

The performance discrepancy between DOMM_RS and DOMM_GT can be reasonably explained by the performance of the RankSVM based ROL prediction subsystem reported in Table 4. Additionally, it is worth noting that arousal AOLs are somewhat unbalanced for IEMOCAP as can be seen from Table 2. The relatively lower number of high and low arousal state could have the effect of increasing the transition probability into the medium arousal state and consequently a higher chance of incorrectly predicting medium [29]. Table 5 also includes the reported performance of the HMM baseline [29], however, as previously mentioned this performance was estimated with different AOLs and data partitions.

Finally, in addition to the summary results in Table 5, we also provide comparisons between the three systems for each individual arousal and valence thresholds ($\theta_{a2}$ and $\theta_{v2}$) employed in our interval label to AOL conversion in Fig. 10 and 11. The red bar on the boxplots indicates the median values across different thresholds and each dot represents performance measured with each threshold. The dashed lines in the boxplots connect dots corresponding to the same threshold and is provided for ease of comparison of systems at the same threshold. It can be seen that the trends observable in Table 5 and even more evident with the additional detail in these figures and specifically DOMM_GT outperforms DOMM_RS, which in turn outperforms OMSVM across every all measures and when using any of the thresholds for interval to AOL conversion.

## 7 CONCLUSION

In the growing body of literature on ordinal emotion prediction, the term "ordinal emotion labels" has been used to refer to two related but different notions. In this paper, we explicitly distinguish between them and introduce the terminology absolute ordinal label (AOL) and relative ordinal label (ROL), conveying the different aspect of ordinality they emobody. Followig this, we establish that the complementary nature of the information encoded in these two different ordinal labelling schemes can be integrated in a joint model of emotion dynamics. Specifically, this novel framework, referred to as the dynamic ordinal Markov model (DOMM), integrates time varying absolute and relative ordinal information.

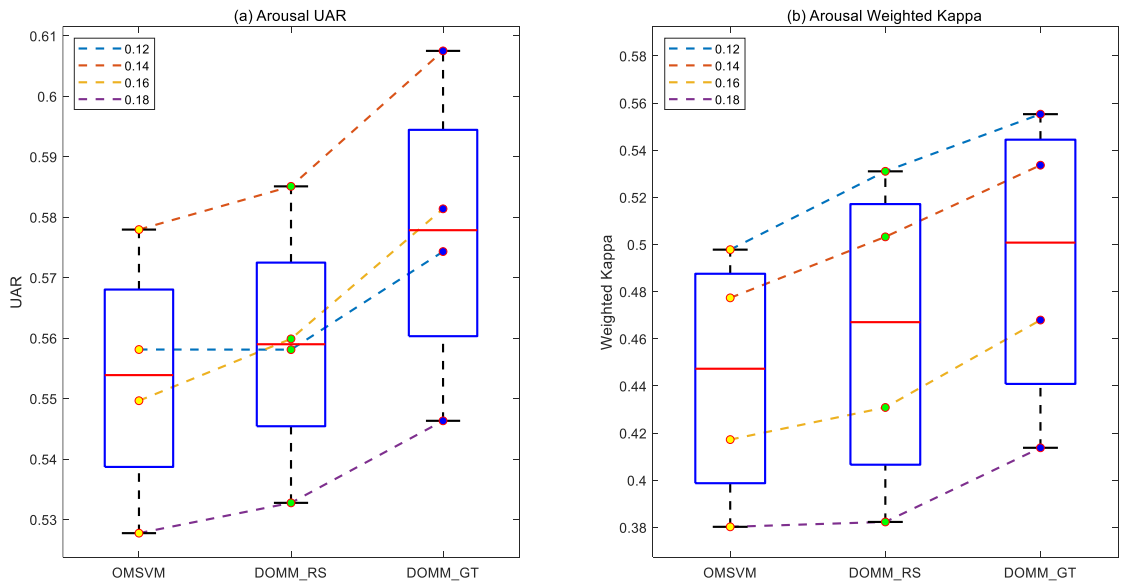

Fig. 10. Unweighted average recall (UAR) and weighted kappa (WK) evaluation of different thresholds on arousal of RECOLA.

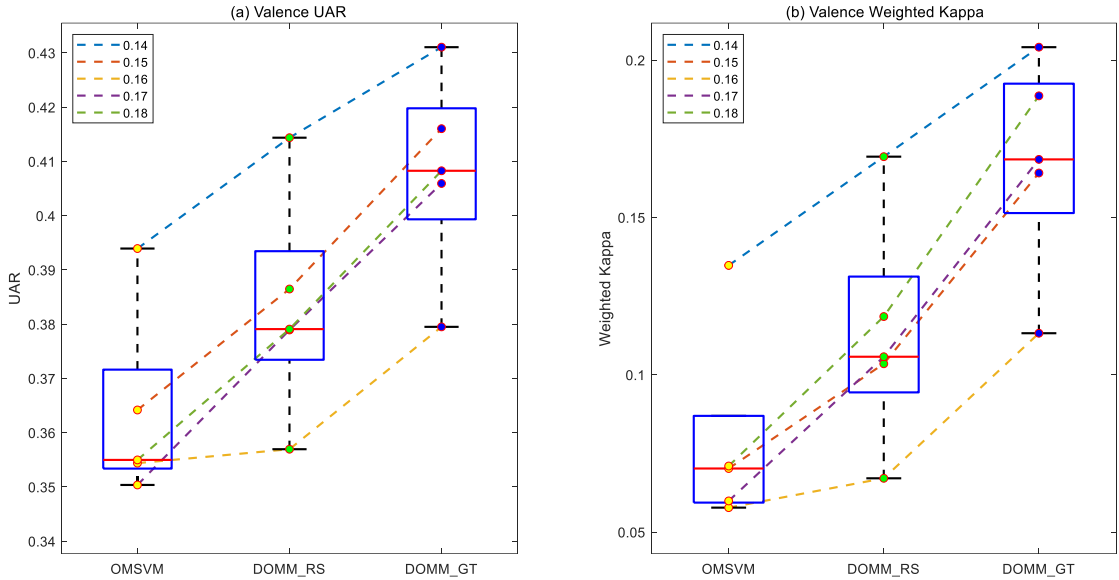

Fig. 11. Unweighted average recall (UAR) and weighted kappa (WK) evaluation of different thresholds on valence of RECOLA.

A realizable and computationally inexpensive, speech-based emotion prediction based on the proposed framework, making use of OMSVM and RankSVM based subsystems for AOL and ROL predictions was implemented to validate the proposed framework. The emotion prediction results obtained across a range of different configurations clearly indicates that integrating relative ordinal information can significantly improve absolute ordinal emotion prediction. The novel system is compared to both, a baseline OMSVM system that only models AOLs; and a version of the novel system that uses oracle ROLs (instead of predicted ROLs) to serve as an indication of the upper bound on the predictive capabilities of the dynamic original Markov model. These experimental comparisons were carried out on both IEMOCAP and RECOLA datasets, and in terms of both UAR and weigthed Kappa. Consistently, in all comparisons, the joint model outperformed AOL only prediction. Finally, the comparison with the version using oracle ROLs suggest that improving the RankSVM based subsystem can lead to further gains in prediction accuracy.

## REFERENCES

[1] V. Sethu, J. Epps, and E. Ambikairajah, "Speech based emotion recognition," in Speech and Audio Processing for Coding, Enhancement and Recognition: Springer, 2015, pp. 197-228.

[2] G. Shashidhar, K. Koolagudi, and R. Sreenivasa, "Emotion recognition from speech: a review," Springer Science+ Business Media, vol. 15, pp. 99-117, 2012.

[3] R. Cowie et al., "Emotion recognition in human-computer interaction," vol. 18, no. 1, pp. 32-80, 2001.

[4] M. El Ayadi, M. S. Kamel, and F. Karray, "Survey on speech emotion recognition: Features, classification schemes, and databases," Pattern Recognition, vol. 44, no. 3, pp. 572-587, 2011.

[5] D. Grandjean, D. Sander, and K. R. Scherer, "Conscious emotional experience emerges as a function of multilevel, appraisal-driven response synchronization," Consciousness and cognition, vol. 17, no. 2, pp. 484-495, 2008.

[6] L. Devillers, L. Vidrascu, and L. Lamel, "Challenges in real-life emotion annotation and machine learning based detection," Neural Networks, vol. 18, no. 4, pp. 407-422, 2005.

[7] R. Plutchik, "Emotions: A general psychoevolutionary theory," Approaches to emotion, vol. 1984, pp. 197-219, 1984.

[8] R. Cowie and R. R. Cornelius, "Describing the emotional states that are expressed in speech," Speech communication, vol. 40, no. 1-2, pp. 5-32, 2003.

[9] J. A. Russell, "A circumplex model of affect," Journal of personality and social psychology, vol. 39, no. 6, p. 1161, 1980.

[10] R. Cowie, C. Cox, J.-C. Martin, A. Batliner, D. Heylen, and K. Karpouzis, "Issues in data labelling," in Emotion-oriented systems: Springer, 2011, pp. 213-241.

[11] G. N. Yannakakis, R. Cowie, and C. Busso, "The ordinal nature of emotions: An emerging approach," IEEE Transactions on Affective Computing, 2018.

[12] A. Metallinou and S. Narayanan, "Annotation and processing of continuous emotional attributes: Challenges and opportunities," in 2013 10th IEEE international conference and workshops on automatic face and gesture recognition (FG), 2013, pp. 1-8: IEEE.

[13] G. A. Miller, "The magical number seven, plus or minus two: Some limits on our capacity for processing information," vol. 63, no. 2, p. 81, 1956.

[14] H. Helson, "Adaptation-level theory: an experimental and systematic approach to behavior," 1964.

[15] J. A. Russell and U. F. Lanius, "Adaptation level and the affective appraisal of environments," vol. 4, no. 2, pp. 119-135, 1984.

[16] G. N. Yannakakis, R. Cowie, and C. Busso, "The ordinal nature

of emotions," in 2017 Seventh International Conference on Affective Computing and Intelligent Interaction (ACII), 2017, pp. 248-255: IEEE.
[17] N. Stewart, G. D. Brown, and N. Chater, "Absolute identification by relative judgment," vol. 112, no. 4, p. 881, 2005.
[18] B. Seymour and S. M. McClure, "Anchors, scales and the relative coding of value in the brain," vol. 18, no. 2, pp. 173-178, 2008.
[19] G. N. Yannakakis and H. P. Martínez, "Ratings are overrated!," Frontiers in ICT, vol. 2, p. 13, 2015.
[20] S. Parthasarathy, R. Lotfian, and C. Busso, "Ranking emotional attributes with deep neural networks," in 2017 IEEE international conference on acoustics, speech and signal processing (ICASSP), 2017, pp. 4995-4999: IEEE.
[21] H. Cao, R. Verma, and A. Nenkova, "Speaker-sensitive emotion recognition via ranking: Studies on acted and spontaneous speech," vol. 29, no. 1, pp. 186-202, 2015.
[22] R. Lotfian and C. Busso, "Practical considerations on the use of preference learning for ranking emotional speech," in 2016 IEEE International Conference on Acoustics, Speech and Signal Processing (ICASSP), 2016, pp. 5205-5209: IEEE.
[23] Y.-H. Yang and H. H. Chen, "Ranking-based emotion recognition for music organization and retrieval," IEEE Transactions on Audio, Speech, and Language Processing, vol. 19, no. 4, pp. 762-774, 2011.
[24] D. Melhart, K. Sfikas, G. Giannakakis, and G. Y. A. Liapis, "A study on affect model validity: Nominal vs ordinal labels," in Workshop on Artificial Intelligence in Affective Computing, 2020, pp. 27-34: PMLR.
[25] H. P. Martinez, G. N. Yannakakis, and J. Hallam, "Don't classify ratings of affect; rank them!," IEEE transactions on affective computing, vol. 5, no. 3, pp. 314-326, 2014.
[26] P. A. Gutierrez, M. Perez-Ortiz, J. Sanchez-Monedero, F. Fernandez-Navarro, and C. Hervas-Martinez, "Ordinal regression methods: survey and experimental study," IEEE Transactions on Knowledge and Data Engineering, vol. 28, no. 1, pp. 127-146, 2016.
[27] I. H. Witten, E. Frank, M. A. Hall, and C. J. Pal, Data Mining: Practical machine learning tools and techniques. Morgan Kaufmann, 2016.
[28] J. Sánchez-Monedero, P. A. Gutiérrez, P. Tiňo, and C. J. N. c. Hervás-Martínez, "Exploitation of pairwise class distances for ordinal classification," vol. 25, no. 9, pp. 2450-2485, 2013.
[29] A. Metallinou, M. Wollmer, A. Katsamanis, F. Eyben, B. Schuller, and S. Narayanan, "Context-sensitive learning for enhanced audiovisual emotion classification," vol. 3, no. 2, pp. 184-198, 2012.
[30] J. C. Kim and M. A. Clements, "Multimodal affect classification at various temporal lengths," vol. 6, no. 4, pp. 371-384, 2015.
[31] M. Neumann, "Cross-lingual and multilingual speech emotion recognition on english and french," in 2018 IEEE International Conference on Acoustics, Speech and Signal Processing (ICASSP), 2018, pp. 5769-5773: IEEE.
[32] Z. Zhang, F. Ringeval, B. Dong, E. Coutinho, E. Marchi, and B. Schüller, "Enhanced semi-supervised learning for multimodal emotion recognition," in 2016 IEEE International Conference on Acoustics, Speech and Signal Processing (ICASSP), 2016, pp. 5185-5189: IEEE.
[33] A. Agresti, Categorical data analysis. John Wiley & Sons, 2003.
[34] J. Verwaeren, W. Waegeman, and B. De Baets, "Learning partial ordinal class memberships with kernel-based proportional odds models," vol. 56, no. 4, pp. 928-942, 2012.
[35] K.-j. Kim and H. Ahn, "A corporate credit rating model using multi-class support vector machines with an ordinal pairwise partitioning approach," Computers & Operations Research, vol. 39, no. 8, pp. 1800-1811, 2012.
[36] H. Ahn and K.-J. Kim, "Corporate credit rating using multiclass classification models with order information," vol. 5, no. 12, pp. 1783-1788, 2011.
[37] J. Kwon, K. Choi, and Y. Suh, "Double Ensemble Approaches to Predicting Firms' Credit Rating," in PACIS, 2013, p. 158: Jeju Island.
[38] L. Cao, L. K. Guan, and Z. Jingqing, "Bond rating using support vector machine," Intelligent Data Analysis, vol. 10, no. 3, pp. 285-296, 2006.
[39] L. R. Rabiner, "A tutorial on hidden Markov models and selected applications in speech recognition," vol. 77, no. 2, pp. 257-286, 1989.
[40] A. W. Bowman and A. Azzalini, Applied smoothing techniques for data analysis: the kernel approach with S-Plus illustrations. OUP Oxford, 1997.
[41] H. Peter D, "Kernel estimation of a distribution function," vol. 14, no. 3, pp. 605-620, 1985.
[42] G. D. Forney, "The viterbi algorithm," vol. 61, no. 3, pp. 268-278, 1973.
[43] N. Fuhr, "Optimum polynomial retrieval functions based on the probability ranking principle," ACM Transactions on Information Systems (TOIS), vol. 7, no. 3, pp. 183-204, 1989.
[44] C. Cortes and V. Vapnik, "Support-vector networks," Machine learning, vol. 20, no. 3, pp. 273-297, 1995.
[45] T. Joachims, "Optimizing search engines using clickthrough data," in Proceedings of the eighth ACM SIGKDD international conference on Knowledge discovery and data mining, 2002, pp. 133-142: ACM.
[46] M. Gonen, A. G. Tanugur, and E. Alpaydin, "Multiclass posterior probability support vector machines," vol. 19, no. 1, pp. 130-139, 2008.
[47] C.-W. Hsu and C.-J. Lin, "A comparison of methods for multiclass support vector machines," IEEE transactions on Neural Networks, vol. 13, no. 2, pp. 415-425, 2002.
[48] A. J. Smola and B. Schölkopf, "A tutorial on support vector regression," Statistics and computing, vol. 14, no. 3, pp. 199-222, 2004.
[49] J. A. Suykens and J. Vandewalle, "Least squares support vector machine classifiers," vol. 9, no. 3, pp. 293-300, 1999.
[50] J. Platt, "Probabilistic outputs for support vector machines and comparisons to regularized likelihood methods," vol. 10, no. 3, pp. 61-74, 1999.
[51] F. Ringeval, A. Sonderegger, J. Sauer, and D. Lalanne, "Introducing the RECOLA multimodal corpus of remote collaborative and affective interactions," in 2013 10th IEEE international conference and workshops on automatic face and gesture recognition (FG), 2013, pp. 1-8: IEEE.
[52] C. Busso et al., "IEMOCAP: Interactive emotional dyadic motion capture database," Language resources and evaluation, vol. 42, no. 4, p. 335, 2008.
[53] M. Valstar et al., "Avec 2016: Depression, mood, and emotion recognition workshop and challenge," in Proceedings of the 6th international workshop on audio/visual emotion challenge, 2016, pp. 3-10.

[54] F. Ringeval et al., "Av+ ec 2015: The first affect recognition challenge bridging across audio, video, and physiological data," in Proceedings of the 5th international workshop on audio/visual emotion challenge, 2015, pp. 3-8.

[55] Z. Huang et al., "An investigation of annotation delay compensation and output-associative fusion for multimodal continuous emotion prediction," in Proceedings of the 5th International Workshop on Audio/Visual Emotion Challenge, 2015, pp. 41-48.

[56] S. Parthasarathy and C. Busso, "Defining Emotionally Salient Regions Using Qualitative Agreement Method," in INTERSPEECH, 2016, pp. 3598-3602.

[57] M. Neumann and N. T. Vu, "Improving speech emotion recognition with unsupervised representation learning on unlabeled speech," in ICASSP 2019-2019 IEEE International Conference on Acoustics, Speech and Signal Processing (ICASSP), 2019, pp. 7390-7394: IEEE.

[58] S. Parthasarathy and C. Busso, "Preference-learning with qualitative agreement for sentence level emotional annotations," in Proc. Interspeech, 2018, pp. 252-256.

[59] O. Verkholyak, D. Fedotov, H. Kaya, Y. Zhang, and A. Karpov, "Hierarchical Two-level modelling of emotional states in spoken dialog systems," in ICASSP 2019-2019 IEEE International Conference on Acoustics, Speech and Signal Processing (ICASSP), 2019, pp. 6700-6704: IEEE.

[60] W. Han, T. Jiang, Y. Li, B. Schuller, and H. Ruan, "Ordinal Learning for Emotion Recognition in Customer Service Calls," in ICASSP 2020-2020 IEEE International Conference on Acoustics, Speech and Signal Processing (ICASSP), 2020, pp. 6494-6498: IEEE.

[61] C.-C. Lee, C. Busso, S. Lee, and S. S. Narayanan, "Modeling mutual influence of interlocutor emotion states in dyadic spoken interactions," in Tenth Annual Conference of the International Speech Communication Association, 2009.

[62] F. Eyben et al., "The Geneva minimalistic acoustic parameter set (GeMAPS) for voice research and affective computing," IEEE Transactions on Affective Computing, vol. 7, no. 2, pp. 190-202, 2016.

[63] L. Tian, J. Moore, and C. Lai, "Recognizing emotions in spoken dialogue with hierarchically fused acoustic and lexical features," in 2016 IEEE Spoken Language Technology Workshop (SLT), 2016, pp. 565-572: IEEE.

[64] F. Eyben, M. Wöllmer, and B. Schuller, "Opensmile: the munich versatile and fast open-source audio feature extractor," in Proceedings of the 18th ACM international conference on Multimedia, 2010, pp. 1459-1462: ACM.

[65] S. Escalera, O. Pujol, and P. Radeva, "Separability of ternary codes for sparse designs of error-correcting output codes," vol. 30, no. 3, pp. 285-297, 2009.

[66] O. Chapelle and S. S. Keerthi, "Efficient algorithms for ranking with SVMs," vol. 13, no. 3, pp. 201-215, 2010.

[67] S. S. Keerthi and D. DeCoste, "A modified finite Newton method for fast solution of large scale linear SVMs," vol. 6, no. Mar, pp. 341-361, 2005.

[68] O. J. N. c. Chapelle, "Training a support vector machine in the primal," vol. 19, no. 5, pp. 1155-1178, 2007.

[69] R.-E. Fan, K.-W. Chang, C.-J. Hsieh, X.-R. Wang, and C.-J. Lin, "LIBLINEAR: A library for large linear classification," vol. 9, pp. 1871-1874, 2008.

[70] J. Cohen, "Weighted kappa: Nominal scale agreement provision for scaled disagreement or partial credit," Psychological bulletin, vol. 70, no. 4, p. 213, 1968.

[71] A. R. Gilpin, "Table for conversion of Kendall's Tau to Spearman's Rho within the context of measures of magnitude of effect for meta-analysis," Educational and psychological measurement, vol. 53, no. 1, pp. 87-92, 1993.

[72] I. StatSoft, "Electronic statistics textbook," Tulsa, OK: StatSoft, 2013.

[73] J. Han, Z. Zhang, F. Ringeval, and B. Schuller, "Reconstruction-error-based learning for continuous emotion recognition in speech," in 2017 IEEE international conference on acoustics, speech and signal processing (ICASSP), 2017, pp. 2367-2371: IEEE.

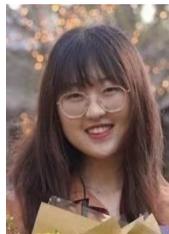

**Jingyao WU** received the BE (Hons) degree in engineering from the University of New South Wales (UNSW), Sydney, Australia in 2020, where she is currently working towards the PhD degree in signal processing. Her research interests include emotion recognition, multimodal signal processing and machine learning. She is a Student Member of the IEEE and SPS.

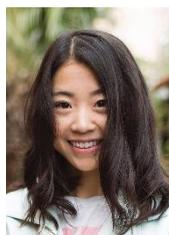

**Ting Dang** received the BE degree and the MEngSc degree in signal processing from Northwestern Polytechnical University, Shaanxi, China, in 2012 and 2015, respectively. She received the PhD degree from the University of New South Wales (UNSW), Sydney, Australia, in 2018. She is currently a postdoctoral research associate in Department of Computer Science, University of Cambridge, UK. Her primary research interests include affective computing, speech processing, audio-based health diagnosis, and machine learning techniques. She is a Member of the IEEE, SPS and ISCA.

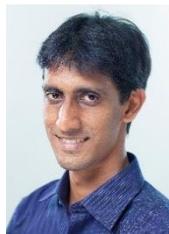

**Vidhyasaharan Sethu** received his B.E. degree from Anna University, Chennai, India, in 2005, his M.Eng.Sc. degree in signal processing, and his Ph.D. degree in speech signal processing from the University of New South Wales (UNSW), Sydney, Australia, in 2006 and 2010, respectively. He is a senior lecturer at the School of Electrical Engineering and Telecommunications, UNSW, Sydney, 2052, Australia. From 2010 to 2013, he was a postdoctoral fellow at the Speech Processing Research Group, UNSW. He has coauthored approximately 100 publications and serves on the editorial board of Computer Speech and Language. His research interests include the application of machine learning to speech processing and affective computing, speaker recognition, and computational paralinguistics. He is a Member of IEEE.

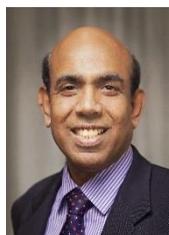

**Eliathamby Ambikairajah** has a B.Sc. (Eng.) (Hons.) degree from University of Sri Lanka and his Ph.D. degree from Keele University, UK. He was the Acting Deputy Vice-Chancellor Enterprise in 2020 at the University of New South Wales, Sydney Australia after previously serving as the Head of School of Electrical Engineering and Telecommunications, from 2009 to 2019. His previous career appointments also include Head of Electronic Engineering and also Dean of Engineering at the Athlone Institute of Technology in Ireland from 1982 to 1999. He was an associate editor of IEEE Transactions on Education from 2012 to 2019. His research interests include speaker and language recognition, emotion detection, machine learning and cochlear modeling. He is a Senior Member of IEEE.